# Self-Organizing Time Map:
# An Abstraction of Temporal Multivariate Patterns

Peter Sarlin[1]

*Åbo Akademi University, Dept. of IT, Turku Centre for Computer Science, Finland*

**Abstract:** This paper adopts and adapts Kohonen's standard Self-Organizing Map (SOM) for exploratory temporal structure analysis. The Self-Organizing Time Map (SOTM) implements SOM-type learning to one-dimensional arrays for individual time units, preserves the orientation with short-term memory and arranges the arrays in an ascending order of time. The two-dimensional representation of the SOTM attempts thus twofold topology preservation, where the horizontal direction preserves time topology and the vertical direction data topology. This enables discovering the occurrence and exploring the properties of temporal structural changes in data. For representing qualities and properties of SOTMs, we adapt measures and visualizations from the standard SOM paradigm, as well as introduce a measure of temporal structural changes. The functioning of the SOTM, and its visualizations and quality and property measures, are illustrated on artificial toy data. The usefulness of the SOTM in a real-world setting is shown on poverty, welfare and development indicators.

**Keywords:** self-organizing time map; self-organizing map; exploratory temporal structure analysis; dynamic visual clustering; exploratory data analysis

---

[1] Corresponding author: Peter Sarlin, Department of Information Technologies, Åbo Akademi University, Turku Centre for Computer Science, Joukahaisenkatu 3–5, 20520 Turku, Finland. Email: psarlin@abo.fi tel. +358 2 215 4670.

# 1. Introduction

During an era of increasing access to complex datasets, abstraction of multivariate temporal patterns is a central issue. However, exploring and extracting patterns in high-dimensional panel data, i.e. along multivariate, temporal and cross-sectional dimensions, is a demanding task. While exploratory data analysis commonly concerns either individual univariate and multivariate time-series or static cross-sectional analysis, a question of central importance is how to combine these tasks. That is, how to identify the occurrence and explore the properties of temporal structural changes in data, as well as their specific locations in the cross section. This type of exploratory data analysis will in the sequel be referred to as exploratory temporal structure analysis.

Kohonen's [1,2] Self-Organizing Map (SOM) is an effective general-purpose tool for abstraction of multivariate mean profiles through projection into a lower dimension. The SOM differs from standard methods for exploratory data analysis by at the same time performing a clustering via vector quantization and projection via neighborhood preservation, as well as by possessing the advantages of a regular grid shape for linking visualizations and a simple and fast learning algorithm. While exploratory analysis with the SOM mainly concerns cross-sectional applications, it is a common tool for classification, clustering and prediction of time-dependent data in a wide range of domains, such as engineering, geographical and environmental sciences, economics and finance (see e.g. [2–4]). The main rationale for using the SOM over more traditional methods for time-series prediction is the inherent local modeling property and topology preservation of units that enhances interpretability of dynamics as well as the availability of growing architectures that facilitate the choice of parsimony (for a thorough review see [5]).

For exploratory analysis on multivariate panel data, however, it is critical to visualize, or present an abstraction across, all dimensions (i.e. multivariate, temporal and cross-sectional spaces). Using a standard two-dimensional SOM for exploratory temporal structure analysis, processing of the time dimension has thus far been proposed along two suboptimal directions: computing separate maps per time unit (e.g. [6–8]) or one map on pooled panel data (e.g. [9–11]). Owing to a possibly high number of time units and temporal differences in correlations and distributions, comparing separate maps per time unit is a laborious task while their structure may not in the least even be comparable. However, SOMs trained with pooled data, for which time can be inferred as a type of latent dimension that is definable but unordered, fail in describing the structure in each cross section. The literature has provided several improvements to the SOM paradigm for temporal processing. We reduce these into four groups: (1) those implicitly introducing time in pre- or post-processing (e.g. trajectories [12]), (2) adaptations of the standard activation and learning rule (e.g. the Hypermap [13]), (3) adaptations of the standard network topology through feedback connections and hierarchical layers (e.g. Temporal SOM [14]) and (4) combinations with other visualization techniques (e.g. interactive spatiotemporal visualization systems [15,16]). Yet, the problem of visualizing changes in inherent data structures over time has not been entirely addressed. The existing SOM literature has thus shortcomings in disentangling the temporal dimensions and cross-sectional structures for exploratory temporal structure analysis, which is the main focus of this paper.

In this paper, we propose a Self-Organizing Time Map (SOTM) for abstraction of the structure in temporal multivariate patterns. In general, the processing of the SOTM is depicted by standard SOM-type learning to one-dimensional arrays for individual time units. We attempt to preserve a stable orientation of the SOTM over time with an initialization based upon short-term memory. When arranging the one-dimensional arrays in an ascending order of time, the SOTM enables a two-dimensional representation with multivariate data structures on the vertical direction and the temporal dimension on the horizontal direction. This output can, when combined with visual aids, be used for dynamic visual cluster analysis, where local distances between SOTM units can be treated as cluster structure information across both directions (i.e. identification of changing, emerging and lost clusters). An ordered SOTM can also be used for projecting individual or grouped data onto the map (constrained by the units of its own time unit). The projections, in conjunction with the structure of the SOTM, enable a temporal version of Bertin's [17] three "levels of reading": elementary level (a view of single multivariate time series), intermediate level (a view of groups of multivariate time series) and global level (a view of temporal multivariate data structures).

For measuring qualities and properties of SOTMs, we adapt several measures and visualizations from the standard SOM paradigm. We also propose a measure for indicating the degree of temporal structural changes in data. A limitation of this work is the absence of a quantitative evaluation, such as commonly performed prediction comparisons to alternative methods. This is, however, due to the lack of a comparable evaluation function. Instead, we illustrate the functioning, output and usefulness of the SOTM on an artificial toy dataset with expected patterns. The generated toy data exhibit multivariate clustered patterns along cross-sectional and temporal dimensions. In addition, we also illustrate drawbacks of a naïve SOM model on these data and provide a guide for interpreting patterns on the SOTM. We also illustrate a real-world application of the SOTM on a temporal multivariate dataset of development and welfare indicators with patterns over the past two decades. The indicators illustrate the progress in fulfilling the Millennium Development Goals (MDGs) – eight goals representing commitments to reduce poverty and hunger and to tackle ill-health, gender inequality, environmental degradation as well as lack of education and access to clean water.

The paper is structured as follows. Section 2 gives an overview of related literature concerning temporal processing with the SOM and attempts to reduce it into four groups. In Section 3, the functioning, visualization and quality and property measures of the SOTM are described. Section 4 illustrates the usefulness of the SOTM, its visualizations and its quality and property measures, as well as a guide for interpreting them, on two datasets: artificial toy data and indicators of the progress towards the MDGs. Section 5 concludes by presenting our key findings and directions for future research.

## 2. Related Work

There is a wide range of literature adapting and extending the standard SOM for temporal processing. While the literature on time in SOMs has been thoroughly reviewed in [6,18–21], a unanimous classification dividing it into distinct groups of studies is far from clear-cut. Drawing upon the above reviews, we attempt to reduce the literature related to the SOTM into four groups of works: those with an implicit consideration of time, those adapting the learning or activation rule, those adapting the topology, and those combining SOMs with other visualization techniques.

The first group applies the standard Kohonen SOM algorithm and illustrates the temporal dimension either as a pre- or post-processing step. The pre-processing concerns embedding a time series into one input vector, such as tapped delay (e.g. [22]). A time-related visualization through post-processing is, however, more common. A connected time series of best-matching units (BMUs), i.e. a trajectory, has been used in the literature to illustrate temporal transitions (e.g. [12,23]). By exploiting the topological ordering of the SOM, visualization of the current and past states enables visual tracking of the process dynamics. In [24,25] , the trajectory approach has been extended to show membership degrees of each time-series point to each cluster. However, while temporal patterns require large datasets for generalization and significance, trajectories can only be visualized for a limited set of data. Thus, strengths and actual directions of the patterns can be obtained by probabilistic modeling of state transitions on the SOM (see e.g. [26,27]).

The second group of works adapts the standard SOM activation or learning rule. Those decomposing the learning rule of the standard SOM into two parts, past and future, for time-series prediction have their basis in the Hypermap [13]. The past part is used for finding best-matching units (BMUs), while the entire input vector is used within the updates of the reference vectors. For predicting out-of-sample data, the past part is again used for finding BMUs while the future part of that unit is the predicted value. This type of learning has been used for standard time-series prediction (see e.g. [28,29]) and predictions through non-linear regression (see e.g. [10,11]). The latter type of decomposition can still be divided into supervised and semi-supervised SOMs, where the difference depends on whether [10] or not [11] the present part is used for matching in training. Instead of considering the context explicitly in SOM training, it can be treated as the neighborhood of the previous BMU. Kangas [30], for instance, constrains the choice of a BMU to the neighborhood of the previous BMU and thus has a behavior that resembles the functioning of SOMs with feedback in the next group.

The third group deals with adaptations of the standard SOM network topology through feedback connections and hierarchical layers. The feedback SOMs have their basis in the seminal Temporal SOM (TSOM) [14] that performs leaky integration to the outputs of the SOM. The recurrent SOM (RSOM) [31–32] differs by moving the leaky integration from the output units to the input vectors. A recent recurrent model is the Merge SOM (MSOM) [33] whose context combines the current pattern with the past by a merged form of the properties of the BMU. The recursive SOM (RecSOM) [34] keeps information by considering the previous activation of the SOM as part of the input to the next time unit, while the feedback SOM (FSOM) [35] differs by integrating an additional leaky loop onto itself. The SOM for structured data (SOMSD) labels, on the other hand, directed acyclic graphs to regular [36] and arbitrary [37] grid structures. Finally, Hammer *et al.* [38] define a general formal framework and show that a large number of SOMs with feedback can be recovered as special cases of it. The hierarchical network architectures, on the other hand, use at each layer one or more SOMs operating at different time scales. The next level in the hierarchy can either use the lower level SOM as input vectors without any processing, such as two-level clustering commonly does, or use transformed input vectors by computing distances between units or concatenating a time series to one input vector, for instance. Kangas [22] introduced hierarchical network architectures to SOMs, and shows that a hierarchical SOM without any additional processing outperforms SOMs with backwards averaged and concatenated input vectors.

The fourth group of studies attempt to create visualization tools for exploratory analysis of spatiotemporal data by combining the SOM with other methods and interactive interfaces. Standard SOMs using both cross-sectional and temporal data have, in addition to trajectory and state-transition analysis, been paired with stand-alone visualization aids for a spatial mapping (e.g. [39]). Guo *et al.* [15] introduces an integrated approach of computational, visual and cartographic methods for visualizing multivariate spatiotemporal patterns, where parallel coordinate plots and reorderable matrices enhance the information products of the SOM. The visualization tool created by Andrienko *et al.* [16] extends the one in [15] by not only grouping spatial situations as per time units, but also spatial locations as per temporal variations. Further, a SOM-based visualization tool for temporal knowledge discovery is introduced in [40,41]. While the tool presents a hierarchical SOM to handle complexity, a U-matrix visualization, trajectory analysis and a transformation of data into linguistic knowledge, it does not, as neither do [15,16], assess temporal changes in data structures.

Indeed, the extensions of all above four groups, while learning temporal transitions and dependencies and mostly providing means for time-series prediction, do not explicitly attempt to and are not directly applicable for exploratory temporal structure analysis. Hence, the comparison of standard two-dimensional SOMs for each time unit, as proposed in [6–8], comes closest to the purpose of the SOTM. Denny *et al.* [7,8] enhance temporal interpretability by applying specific initializations and visualizations. Nevertheless, the method has the drawback of an unstable orientation over time and complex comparisons of two-dimensional grids. Hence, we propose here a Self-Organizing Time Map (SOTM) for a two-dimensional abstraction of temporal multivariate patterns with capabilities for conducting exploratory temporal structure analysis on a large number of time units.

## 3. Self-Organizing Time Map and its Properties, Qualities and Visualizations

Kohonen's [1,2] SOM paradigm is based upon projection via preservation of neighborhood relations and clustering via vector quantization. The SOM performs a mapping from the input data space $\Omega$ onto a $k$-dimensional array of output units by attempting to preserve the neighborhood relations in data. The vector quantization capability of the SOM allows modeling from the continuous space $\Omega$, with a probability density function $p(x)$, to the array of units, whose location depend on the neighborhood structure of $\Omega$. When $k=2$, as is common in exploratory data analysis, the positions on the horizontal and vertical axes do not carry a parametric meaning; they represent positions of mean profiles in the data space.

### 3.1 Self-Organizing Time Map (SOTM)

The Self-Organizing Time Map (SOTM) uses the clustering and projection capabilities of the standard SOM for visualization and abstraction of temporal structural changes in data. However, here $t$ (where $t=1,2,\ldots,T$) is a time-coordinate in data, not in training iterations as is common for the standard SOM. To observe the cross-sectional structures of the dataset for each time unit $t$, the SOTM performs a mapping from the input data space $\Omega(t)$, with a probability density function $p(x,t)$, onto a one-dimensional array $A(t)$ of output units $m_i(t)$ (where $i=1,2,\ldots,M$). After performing a mapping for all $t$, the timeline is created by arranging $A(t)$ in an ascending order of time $t$. The positions on the SOTM carry a different meaning than those on the standard SOM; the horizontal direction has a parametric interpretation of time $t$ while the vertical direction represents positions in the data space $\Omega(t)$. Hence, the topology is rectangular rather than hexagonal and topology preservation is twofold, where the horizontal direction preserves time topology and the vertical preserves data topology.

The orientation preservation and gradual adjustment to temporal changes is performed as follows. The first principal component of principal component analysis (PCA) is used for initializing $A(t_1)$ and setting the orientation of the SOTM. For preserving the orientation between consecutive patterns in a time series, the model uses short-term memory to retain information about past patterns. Thus, the orientation of the map is preserved by initializing $A(t_{2,3,\ldots,T})$ with the reference vectors of $A(t-1)$. Adjustment to temporal changes is achieved by performing a batch update per time $t$. For $A(t)$ (where $t=1,2,\ldots,T$), each data point $x_j(t) \in \Omega(t)$ (where $j=1,2,\ldots,N(t)$) is compared to reference vectors $m_i(t) \in A(t)$ and assigned to its best-matching unit (BMU) $m_c(t)$:

$$\|x(t) - m_c(t)\| = \min_i \|x(t) - m_i(t)\|.$$

Then each reference vector $m_i(t)$ is adjusted using the batch update formula:

$$m_i(t) = \frac{\sum_{j=1}^{N(t)} h_{ic(j)}(t) x_j(t)}{\sum_{j=1}^{N(t)} h_{ic(j)}(t)},$$

where index $j$ indicates the input data that belong to unit $c$ and the neighborhood function $h_{ic(j)}(t) \in (0,1]$ is defined as a Gaussian function

$$h_{ic(j)}(t) = \exp\left(-\frac{\|r_c(t) - r_i(t)\|^2}{2\sigma^2}\right),$$

where $\|r_c(t) - r_i(t)\|^2$ is the squared Euclidean distance between the coordinates of the reference vectors $m_c(t)$ and $m_i(t)$ on the one-dimensional array, and $\sigma$ is the user-specified neighborhood parameter. From this follows obviously that neighborhood $\sigma$ only includes vertical relationships. In contrast to what is common for the standard batch SOM, the neighborhood $\sigma$ is constant over time for a comparable timeline, not a decreasing function of time as is common when time represents iterations.

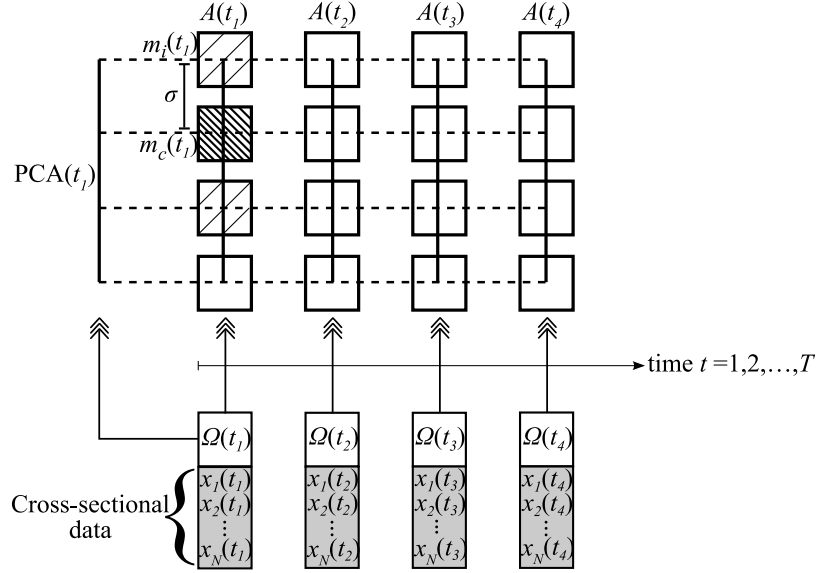

Figure 1. The functioning principles of the SOTM

To sum up, Figure 1 presents the functioning principles of the SOTM, which can be distinguished as follows:

    *t* = 1
    initialize *A*(*t*) using PCA on *Ω*(*t*)
    apply the batch update to *A*(*t*) using *Ω*(*t*)
    while *t*<*T*
       *t* = *t* + 1
       initialize *A*(*t*) using the reference vectors of *A*(*t*-1)
       apply the batch update to *A*(*t*) using *Ω*(*t*)
    end
    order *A*(*t*) in an ascending order of time *t*

### 3.2 SOTM Properties

The above presented SOTM specification, while being flexible in nature, disposes some assumptions on distance metrics and grid shapes, as well as other computational details. Even though a SOTM mapping to one-dimensional arrays looses in granularity and detail to the two-dimensional, the sole case of successful complete mathematical study of the SOM is in one dimension (though with one input dimension as well) (for a review see [42]). Further, a two-dimensional representation, while comprising of less details, facilitates interpretation over the three-dimensional case. The SOTM is implemented using the Euclidean metric for the sake of simplicity and purpose herein as well as stick to the standard batch SOM with exponential neighborhood functions. We prefer the batch over the sequential SOM for its well-known properties of efficiency and precision (see e.g. [1]). When compared in terms of computational cost, the SOTM is cheaper than a standard SOM of the same size since matching and learning is restricted by time. The SOTM also has the asset of keeping the most important properties and the interpretation of the SOM as it has its basis in the very standard SOM algorithm. As the SOTM processes data in batches per time *t*, the disadvantage of all data points having to be available when training begins is not a concern given that all data for each *t* are accessible simultaneously. In this sense, the SOTM can be seen as a type of online batch SOM. While the purpose of use of the SOTM is different, the functioning of it can also be linked to several other pieces of literature extending the SOM. For instance, the increase in number of units over time resembles the functioning of Growing SOMs [43] and the short-term memory initialization resembles SOMs with feedback connections (e.g. [34]). While the SOTM herein uses specifications from the very standard SOM literature, such as batch training, Euclidean metric and exponential neighborhoods, its matching, learning and neighborhoods could be implemented in various modified fashions, such as those discussed in the related literature (Section 2). Parameterization of batch training can also be performed in a number of ways depending on the task and data at hand. For instance, the first array $A(t_1)$ may be trained until convergence if the initialization is far from converged and the number of training cycles of each array $A(t)$ may be increased if quantization accuracy is relatively important. Idle units, i.e. units not attracting any data, while representing a discrepancy between the initialized array $A(t)$ and data $\Omega(t)$, may also be dealt with through increases in training cycles.

### 3.3 Quality and Property Measures of the SOTM

Common quality measures for evaluating the goodness of a SOM are quantization error, distortion measure and topographic error. These, as well as other measures of the SOM, could be adapted to apply for quantifying the qualities and properties of SOTMs, where quality refers to the goodness of the mapping and property to characteristics of the data. Computations of quality and property measures can be distinguished as follows: quality measures of SOTMs are summed over *T* whereas property measures of a SOTM depict the characteristics of data at each *t*. However, property measures obviously also illustrate time-specific qualities of SOTMs. The fit of the SOTM to the data distribution can be measured with an adaptation of the standard quantization error and distortion measure. The time-restricted quantization errors $\varepsilon_{qe}$ and $\varepsilon_{qe}(t)$ compute the average distance between $x_j(t) \in \Omega(t)$ and $m_c(t) \in A(t)$:

$$\varepsilon_{qe} = \frac{1}{T}\sum_{t=1}^{T}\frac{1}{N(t)}\sum_{j=1}^{N(t)}\left\|x_j(t) - m_{c(j)}(t)\right\|$$

$$\varepsilon_{qe}(t) = \frac{1}{N(t)}\sum_{j=1}^{N(t)}\left\|x_j(t) - m_{c(j)}(t)\right\|.$$

The distortion measures $\varepsilon_{dm}$ and $\varepsilon_{dm}(t)$ indicate similarly the fit of the map to the shape of the data distribution, but also account for the radius of the neighborhood:

$$\varepsilon_{dm} = \frac{1}{T}\sum_{t=1}^{T}\frac{1}{N(t)}\frac{1}{M(t)}\sum_{j=1}^{N(t)}\sum_{i=1}^{M(t)}h_{ic(j)}(t)\left\|x(t)_j - m(t)_i\right\|^2$$

$$\varepsilon_{dm}(t) = \frac{1}{N(t)}\frac{1}{M(t)}\sum_{j=1}^{N(t)}\sum_{i=1}^{M(t)}h_{ic(j)}(t)\left\|x(t)_j - m(t)_i\right\|^2.$$

The topology preservation of the SOTM can also be measured using an adaptation of the standard topographic error. The time-restricted topographic errors $\varepsilon_{te}$ and $\varepsilon_{te}(t)$ measure by $u(x_j(t))$ the average proportion of $x_j(t) \in \Omega(t)$ for which first and second BMUs (within $A(t)$) are non-adjacent units:

$$\varepsilon_{te} = \frac{1}{T}\sum_{t=1}^{T}\frac{1}{N(t)}\sum_{j=1}^{N(t)}u(x_j(t))$$

$$\varepsilon_{te}(t) = \frac{1}{N(t)} \sum_{j=1}^{N(t)} u(x_j(t)).$$

While quantifying the degree of temporal changes in data is of central importance, it is oftentimes a difficult task. The SOTM enables approximating the structural change between time units *t*-1 and *t* by an average Euclidean distance between $m_i(t-1) \in A(t-1)$ and $m_i(t) \in A(t)$ for all pairs *i*=1,2,…,*M*. The distance is meaningful given that the ending point of *A*(*t*-1) is the starting point of *A*(*t*) and as the adjustment to temporal changes ($\sigma$) is constant over time. The structural changes $\varepsilon_{sc}$ and $\varepsilon_{sc}(t)$ are computed as follows:

$$\varepsilon_{sc} = \frac{1}{T} \sum_{t=1}^{T} \frac{1}{M} \sum_{i=1}^{M} \|m_i(t-1) - m_i(t)\|$$

$$\varepsilon_{sc}(t) = \frac{1}{M} \sum_{i=1}^{M} \|m_i(t-1) - m_i(t)\|.$$

When the quantization error $\varepsilon_{qe}(t)$, distortion measure $\varepsilon_{dm}(t)$ and topographic error $\varepsilon_{te}(t)$ are computed for *t*=1,2,…,*T* and structural change $\varepsilon_{sc}(t)$ for *t*=2,3,…,*T*, they can be plotted over time. This is useful for identifying properties and qualities of data at each time unit, in particular the degree of temporal changes in data.

### 3.4 Visualizations of the SOTM

The output of the SOTM is a two-dimensional array of units, with time on the horizontal direction and data structures on the vertical, which represents a multidimensional space. While there exist numerous visualizations for the SOM that could be applied to the SOTM framework, we present here the standard ones (see e.g. [44]) which enhance the objectives of the SOTM. The multidimensionality of the array can be visualized through layers, or feature planes. For each individual input, a feature plane represents the spread of its values. Thus, one can interpret vertical differences as cross-sectional properties and horizontal differences as temporal changes. As for the standard SOM, the feature planes are different views of the same map, where one unique point represents the same unit on all planes. The coloring of the feature planes is here performed using a ColorBrewer's [45] scale, where variation of a blue hue occurs in luminance and light to dark represent low to high values according to a feature plane-specific scale. As the scale is common for the entire SOTM (i.e. *A*(*t*) where *t*=1,2,…,*T*) for each feature plane, the temporal changes in the spread of values are shown by variations in shade. While pre-processing of data is necessary for standardized weighting of input variables, we facilitate the interpretation of the feature planes by post-processing the variables back to their original distributions.

While plots of $\varepsilon_{qe}(t)$, $\varepsilon_{dm}(t)$, $\varepsilon_{te}(t)$ and $\varepsilon_{sc}(t)$ show the changes over time, the assessment of structural differences on the horizontal and vertical dimensions of the SOTM can be enhanced by a Multidimensional scaling (MDS) method, such as Sammon's non-linear mapping [46]. The reason for preferring Sammon's mapping over other MDS methods is its focus on local distances. Time is disentangled by mapping all multidimensional SOTM units $m_i(t)$ (where *t*=1,2,…,*T*) to one dimension using Sammon's mapping and then plotting that dimension individually for each time *t*. Thus, this representation has Sammon's dimension on the *y* axis and time on the *x* axis. By connecting adjacent units with solid (data topology) and dashed (time topology) lines for a net-like representation and showing topographic errors $u(x_j(t))$ through color coding, we facilitate the detection of structural changes and topographic errors. Moreover, a coloring method based upon that in Kaski *et al.* [39] for revealing changes in cluster structures can be applied to the SOTM. The well-known uniform color space CIELab [47] is used, where perceptual differences of colors represent distances in the data space, as approximated by the Sammon's mapping. However, as the SOMs of the SOTMs are one-dimensional, we only use one dimension (blue to yellow) of the color space. Indeed, other techniques for SOM visualization can, and should, be adapted for enhancing the SOTM visualization, such as the common Unified distance matrix (U-matrix) [48] and a spatial mapping of cluster coloring [39], a second-level clustering, and Aupetit's proximity measure [49].

**4. Experiments**

We illustrate the functioning, output and quality and property measures of the SOTM on two datasets: an artificial toy dataset and a dataset with welfare and development indicators. The former attempts to validate the use of a SOTM over a naïve SOM model and the output of the SOTM by representing expected patterns, as well as provides a guide for interpreting patterns on a SOTM. The latter application illustrates the use of the SOTM in a real-world setting.

**4.1 Illustrative Toy Examples**

To validate the performance of the SOTM, we generate an artificial toy dataset with expected patterns. The data need to come from a three-dimensional cube, where one dimension represents time, one the cross-sectional entities and one the input variables. The toy data are generated by setting five weights $w_{1-5}$ that adjust a mixture of randomized shocks on four different levels: group-specific ($g$), time-specific ($t$), variable-specific ($r$) and common ($j$) properties. For each variable, group-level differences are included to have artificial clusters, time-level properties to introduce temporal trends, and group-specific and common shocks to introduce general noise. We generate data $x(r,g,j,t)$ by combining group-specific trends $E$ with shocks across data and over time,

$$x(r,g,j,t) = \frac{1}{1+\exp(-[E(r,g,t)+w_4(r,g)e_4(r,t)+w_5(r,g)e_5(r,j,t)])}$$

and

$$E(r,g,t) = w_1(r)e_1(g) + w_2(r)e_2(g)t + w_3(r)e_3(g,t),$$

where $e_{1,3-5} \sim N(0,1)$, $e_2 \sim U(0,1)$, $r$ stands for variables, $g$ for groups, $t$ for time and $j$ for entities, and $E$ computes group-specific trends. Weights $w_{1-5}$ specify the following properties of data: $w_1$ sets the group-specific intercepts, $w_2$ the group-specific slopes over time, $w_3$ the magnitude of group-specific random shocks, $w_4$ the magnitude of time-specific common shocks and $w_5$ the magnitude of common shocks. Figure 2 plots the four variables and reports the used weights $w_{1-5}$ for generating 100 entities over 10 periods, where the color coding illustrates five entity groups. Particular characteristics of the below four variables are as follows: $x_1$ has small differences in intercepts and a positive slope; $x_2$ has large differences in intercepts, a negative slope and minor group-level and common shocks over time and across entities; $x_3$ has large differences in intercepts, and a constant trend with minor common shocks across entities and over time; and $x_4$ has large differences in intercepts and large common shocks over time.

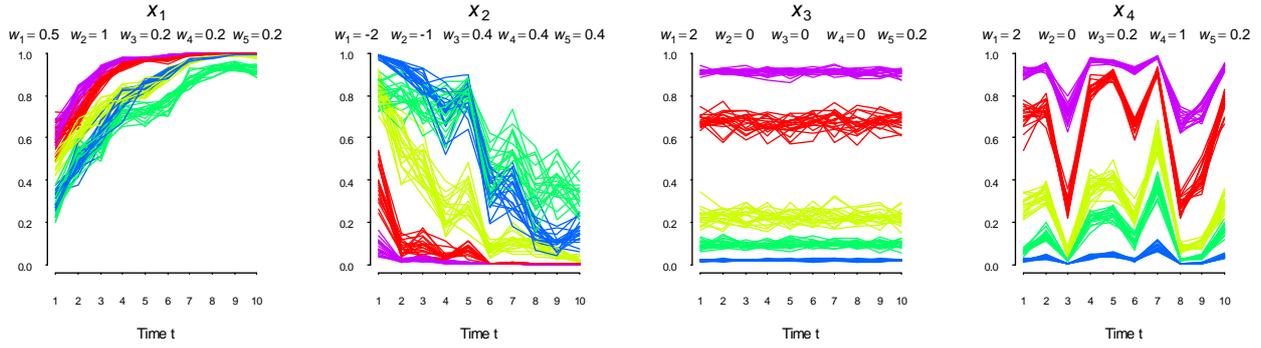

Figure 2. Plots of the four generated toy variables $x_{1-4}$ with weights $w_{1-5}$. Data consist of 5 groups of 20 cross-sectional entities over 10 periods, where the color coding illustrates the groups, the *x* axis represents time and *y* axis the values.

### 4.1.1 A Naïve SOM Model

Comparing methods for exploratory data analysis is not an entirely straightforward task. The absence of a quantitative evaluation, such as common prediction or classification comparisons, is due to the lack of a comparable evaluation function. Instead, we focus herein on illustrating the advantages of the SOTM by comparing it to a naïve one-dimensional SOM model on the entire dataset $\Omega$. While being somewhat trivial, the exercise attempts to illustrate how time, when being embedded, cannot be fully represented on a standard SOM, not even when utilizing post-processing techniques. In this SOM, the pooled toy dataset is used as an input to a SOM with 5 units as per the number of groups in data. Figure 3 shows the SOM, its feature planes and a post-processed trajectory for the toy dataset.

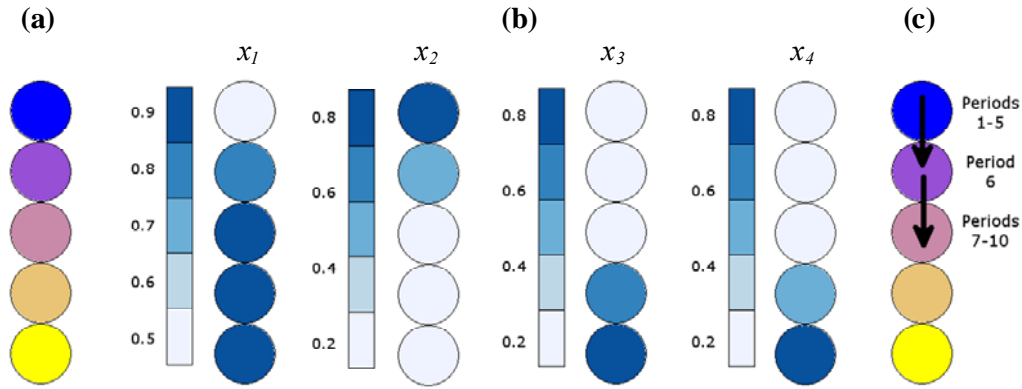

Figure 3. A naïve one-dimensional SOM (a), layers or feature planes of the SOM (b) and an exemplification of temporal movements of an arbitrary data point on the SOM (c).

Figure 3a shows the SOM where differences in units are represented by perceptual differences in colors. Its feature planes in Figure 3b depict characteristics of the data in Figure 2, but obviously disregard the time dimension. For instance, neither time trends of $x_1$ and $x_2$ nor time shocks of $x_4$ are depicted. Variable $x_3$ is, however, correctly depicted as it is close to constant over time. A trajectory of an entity can be used for describing its evolution on the SOM over time. In Figure 3, a trajectory of an arbitrary data point over the 10 periods exemplifies that, while temporal movements of individual data exist, changes in cluster (or unit) structures are not represented. In particular, this illustrates that the evolution of data structures in Figure 2 is not represented by a static SOM.

### 4.1.2 A Toy Example of the SOTM

Next, we apply the SOTM on the toy data. While we use the standard SOTM specification presented in Section 3, we still have to set the architecture of it. The SOTM is chosen to have 5x10 units, where 5 units represent data topology at time *t* on the vertical direction and 10 units the time topology on the horizontal direction. The number of units on the horizontal axis is set by the number of time units *T* in data, while the number of units (or clusters) on the vertical axis equals the number of groups in data. The quality measures presented in Section 3 are used for evaluating performance over different parameters. For all time units *t*, the distortion measure $\varepsilon_{dm}$ and quantization error $\varepsilon_{qe}$ measure the fit to data $\Omega$, while topographic error $\varepsilon_{te}$ measures the aggregated topology preservation. The structural change $\varepsilon_{sc}$, on the other hand, shows the distance between horizontal units. Figure 4 shows the quality measures over radius of the neighborhood $\sigma$ ranging from 0.4 to 8. The figure illustrates aspects of not only these data in particular, but also SOTM training in general. It shows the strength of the topology preservation in the SOTM; a topology error $\varepsilon_{te}$ is only found for experiments with $\sigma = 0.4$. Though the magnitude of quantization error $\varepsilon_{qe}$ and distortion measure $\varepsilon_{dm}$ differs due to simple and squared distances, an obvious effect is the increase of the measures when $\sigma$ increases. The structural change $\varepsilon_{sc}$ starts to decrease when $\sigma = 0.6$, and decreases until it stabilizes for $\sigma \geq 1.6$. When aiming at data abstraction and exploratory analysis, choosing optimal parameter values for a SOTM, likewise for a SOM, is a difficult task; the choice can be said to depend on the relative preferences of the analyst between topographic and quantization errors. However, as the interpretation of a SOTM relies heavily on topology preservation, not least the time dimension, topographic error ought to be of higher importance. As we here only have topographic errors for $\sigma = 0.4$, we can choose a SOTM with minimum quantization error and distortion. The chosen SOTM has thus a radius of the neighborhood $\sigma = 1.6$.

The final SOTM is found in Figure 5a and a Sammon's mapping of it in Figure 5b. The coloring of the SOTM uses the CIELab unified color space, where perceptual differences in colors represent differences between units as approximated by Sammon's mapping. Feature planes in Figure 5c represent layers of the SOTM, while Figure 5d reports trajectories of all data on the SOTM. Figure 7 illustrates a plot of property measures $\varepsilon_{qe}(t)$, $\varepsilon_{dm}(t)$, $\varepsilon_{te}(t)$ and $\varepsilon_{sc}(t)$ over time.

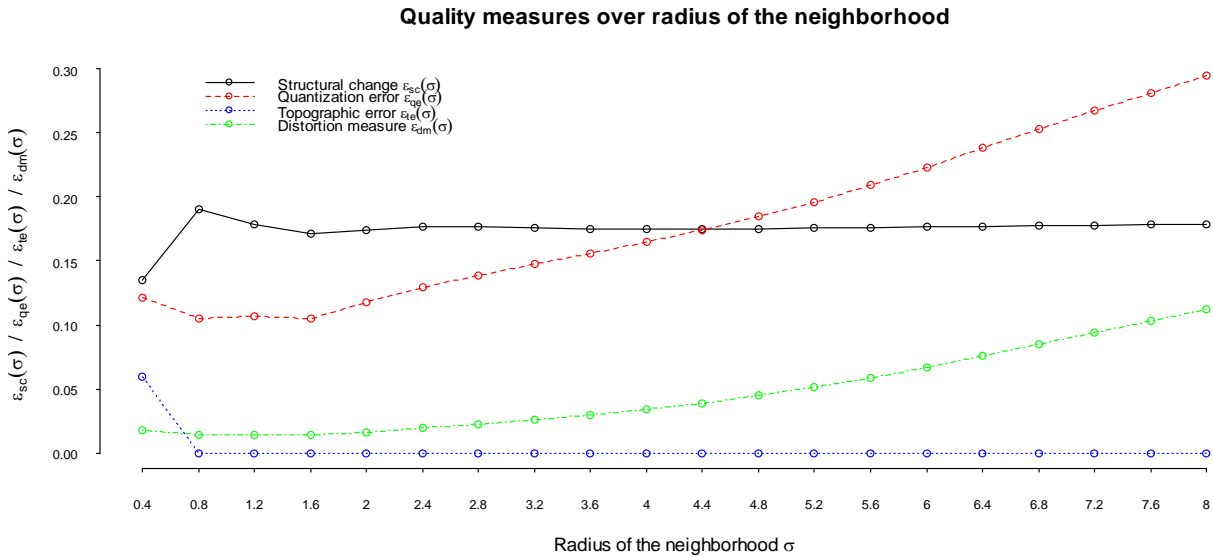

Figure 4. Quality measures over radius of the neighborhood on the toy dataset. For models with a 5x10 array of units, the errors ($\varepsilon_{qe}$, $\varepsilon_{dm}$, and $\varepsilon_{te}$) are computed as aggregates of all time units *t*=1,2,…,*T* and structural changes $\varepsilon_{sc}(t)$ of time units *t*=2,3,…,*T* over neighborhood radii $\sigma = \{0.4, 0.8, ..., 8\}$.

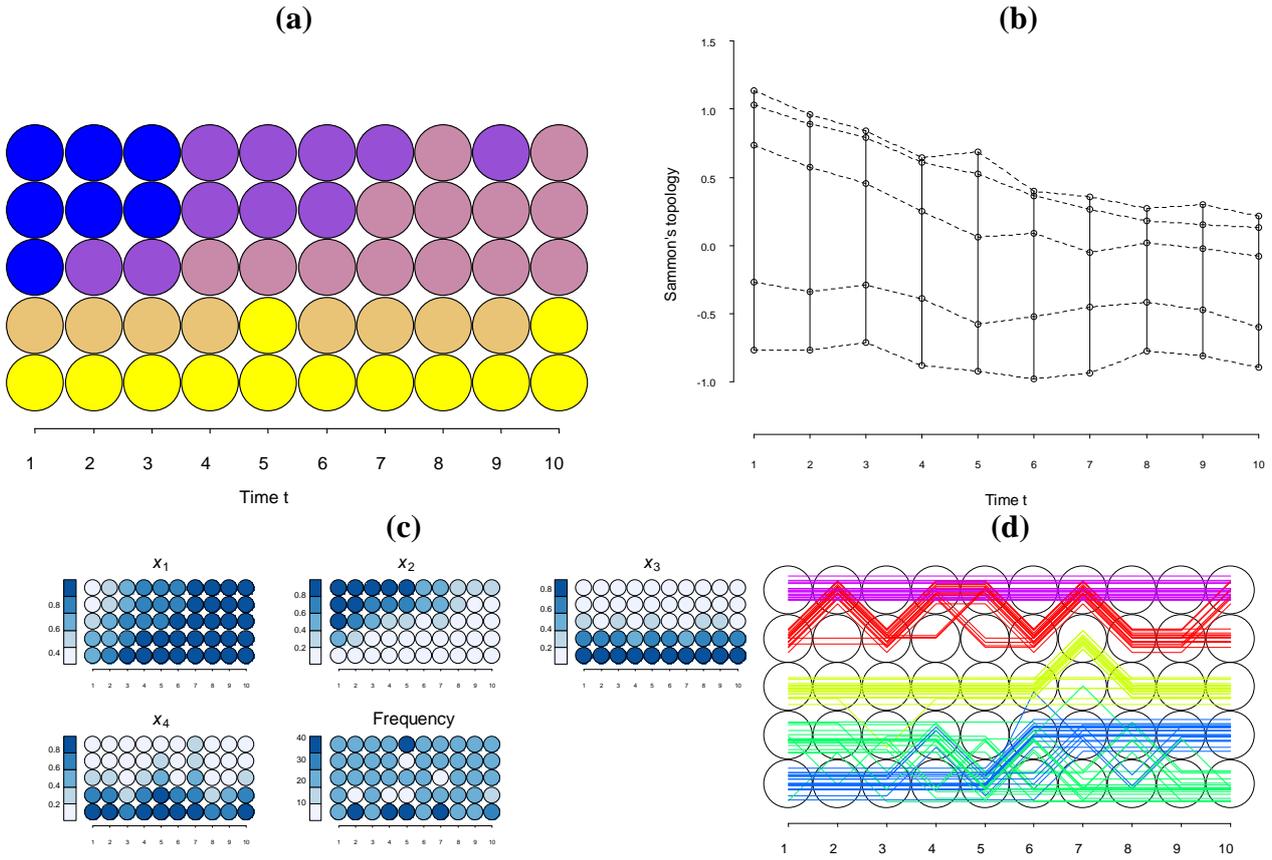

Figure 5. A SOTM grid with perceptual differences in color representing distances between units (a), a plot of the SOTM units according to Sammon's topology on the vertical axis and time on the horizontal axis where neighboring units are connected with lines (b), individual input layers, or feature planes, as well as a frequency plot on the SOTM grid (c), and the data overlaid as time series, or trajectories, on top of the SOTM grid with coloring that corresponds to that in Figure 2 (d).

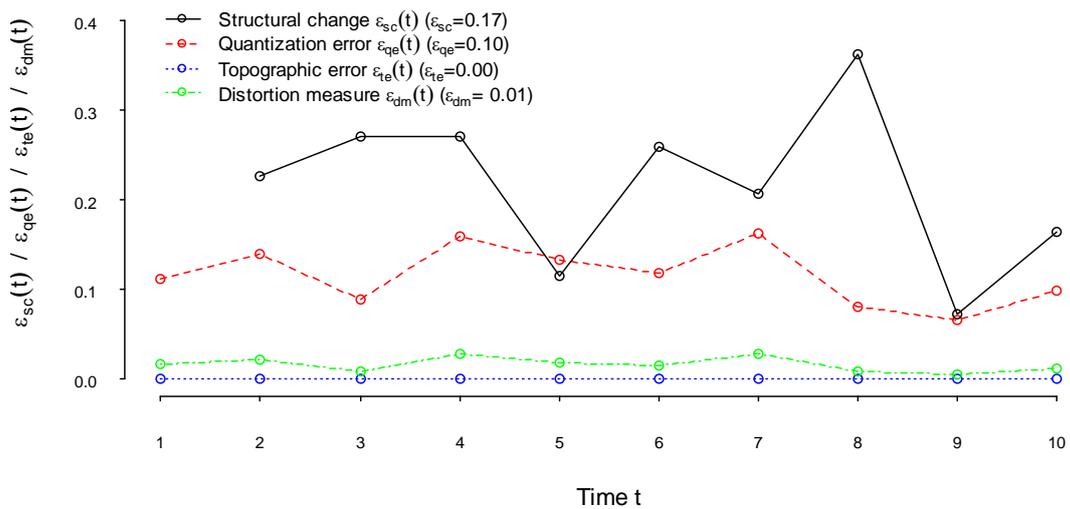

Figure 6. Property measures of the SOTM. The errors ($\varepsilon_{qe}(t)$, $\varepsilon_{dm}(t)$, and $\varepsilon_{te}(t)$) are computed for time units $t=1,2,\ldots,T$ and $\varepsilon_{sc}(t)$ for time units $t=2,3,\ldots,T$ for the final model with an 5x10 array of units and $\sigma = 1.6$.

### 4.1.3 A Guide for Interpreting the SOTM

In this section, we will give a brief guide for interpreting the SOTM and its visualizations. A key to interpreting the SOTM is to understand the grid structure and the following representation of data along two directions. The vertical direction (or columns of units or axis) has a similar interpretation as a standard SOM (cf. Figure 3), but each one refers to a specific time unit. Thus, it represents the cross-sectional data structure, or data topology, at time $t$, where similar units are located close together. The horizontal direction (or rows of units or axis), while being conceptually different from a standard SOM, has a similar interpretation. It represents the time structure, or time topology, where similar units are again located close together, but refers instead to resembling units at different points in time. Hence, differences along both directions represent differences between respective topologies when interpreting properties of high-dimensional structures, values of individual inputs or any other linked information. Below we will use the above toy example for illustrating the interpretation of the SOTM visualizations in this paper.

Figures 5a and 5b give information on the distance structure of the SOTM. Perceptual differences in colors (blue to yellow) in Figure 5a represent differences between units as per distances in the Sammon's mapping in Figure 5b. In Figure 5b, differences between units on both vertical and horizontal directions should, however, be interpreted by values of Sammon's topology (color in Figure 5a and $y$ axis in Figure 5b). The differences in values of units on the vertical direction represent distances in cross-sectional data structures at a specific time $t$ and differences in values of units on the horizontal direction represent distances over time. In the Sammon's mapping, solid connections between units represent data topology and dashed connections time topology. The figures show that data are clustered into two distinct groups: the three uppermost horizontal rows (yellow, green and blue, cf. Figure 2) and two lowest rows (red and purple, cf. Figure 2). The structure of the SOTM illustrates two types of temporal changes: common trends of the entire structures and movements of individual units. The former type preserves distances between units at each point in time, but moves the entire structure to some direction, while the latter type illustrates changes in distances to neighboring units, i.e. identification of changing, emerging and lost clusters. Figures 5a and 5b show that the two distinct groups converge over time, in particular that the uppermost groups of data move towards the rest of the data, as the raw data in Figure 2 confirm. Convergence is mostly a result of inputs $x_1$ and $x_2$ moving towards maximum and minimum values over time, in particular the large changes of $x_2$.

Figure 5c illustrates the spread of values for each of the four inputs and should similarly be interpreted along the two directions. One type of validation of the SOTM is that the four feature planes correspond to the description of differences in group-level intercepts and slopes, as well as time-specific shocks, for the inputs (cf. Figure 2). For instance, $x_1$ has small differences in intercepts and a positive slope, $x_2$ has large differences in intercepts and a negative slope, $x_3$ has large differences in intercepts and a constant trend, and $x_4$ has large differences in intercepts and large common shocks over time. The frequency plane in Figure 5c represents density of data on the SOTM grid and is particularly useful for two purposes. Since the SOTM attempts to update cluster structures in $A(t-1)$ to $A(t)$ by a batch update, while structures in data $\Omega(t-1)$ and $\Omega(t)$ may be of different nature, one purpose of use is locating idle units. While idle units represent a change in cluster structures, the reference vectors are still transmitted to $A(t)$ through the short-term memory.[2] The frequency plots also enable observing evolution of densities over time. While changes in the spread of values indeed indicate changes in data, frequencies are an equally important property of data. In this toy example, the main interpretation is the absence of idle units. An even better validation of the SOTM than the feature planes is the plot of all individual data on the SOTM in

---

[2] We suggest to deal with idle units through some color coding. While we have implemented each idle unit to be colored as grey, we do not encounter these specific cases in the experiments performed in this paper.

Figure 5d. The coloring of the trajectories corresponds to that in Figure 2 and illustrates the evolution of the groups on the SOTM. While the groups are separated during most of the periods, some overlap and interchange of positions occurs over time. The one-period overlaps of red and purple groups accurately correspond to the time-specific shocks of $x_4$. The occurrence of position interchanges of blue and green groups at periods 3-5 are likely due to change in input $x_1$ and finally in period 7 due to substantial changes in input $x_2$.

Plots of property measures over time in Figure 6 illustrate the variation of $\varepsilon_{qe}(t)$, $\varepsilon_{dm}(t)$, $\varepsilon_{te}(t)$ and $\varepsilon_{sc}(t)$ over time. When assessing properties for each time unit $t$, the structural change $\varepsilon_{sc}(t)$ measures divergence of $m_i(t)$ from the units $m_i(t-1)$, whereas the rest mainly visualize quantization and topographic qualities across a SOTM. While increases in quantization error $\varepsilon_{qe}(t)$ and distortion $\varepsilon_{dm}(t)$ represent the fit of data $\Omega(t)$ to units $m_i(t)$, increases in topographic error represents the topology preservation for each array $A(t)$. For the toy data, the large variation in $\varepsilon_{sc}(t)$ depicts the existence of large differences between data structures. In particular, we can see that highest values of $\varepsilon_{sc}(t)$ in periods 3-4 and 6-8 co-occur with large common temporal shocks in $x_4$. Small or none variation of $\varepsilon_{qe}(t)$, $\varepsilon_{dm}(t)$ and $\varepsilon_{te}(t)$ confirms that quantization and topographic errors are low over time, while the difference in the magnitude of the quantization accuracies is a result of them being measured with simple and squared distances, respectively.

**4.2 Millennium Development Goals**

This real-world application presents an abstraction of a selection of World Development Indicators for tracking the progress of the MDGs with patterns during 1990–2008. World Bank's database on World Development Indicators has commonly been used for demonstrating SOM processing and its extensions (e.g. [7,8,39,50,51]). The eight MDGs represent commitments to reduce poverty and hunger, and to tackle ill-health, gender inequality, and lack of education and access to clean water as well as environmental degradation by 2015. The dataset consists of yearly matrices $\Omega(t)$ for $t=1,2,\ldots,19$, where rows represent countries and columns represent 15 indicators measuring fulfillment of the MDGs. The dataset consists of 207 countries spanning from 1990–2008. For standardized weighting of the inputs, we transform each indicator by variance to have mean 0 and standard deviation 1, but obviously do it on distributions spanning the entire dataset $\Omega$.

When training SOTMs and choosing the final specification, we measure performance using the criteria presented in Section 3. The architecture of the SOTM is chosen to be 8x19 units, where eight units represent data topology at time $t$ and 19 the time topology. The number of units (or clusters) on the vertical axis equals the number of clusters found in a similar pooled dataset $\Omega$ in [50]. However, the SOTM, likewise the SOM, is not restricted to treat each unit as an individual cluster. Due to the property of approximating probability density functions $p(x)$, sparse locations tend not to, while dense tend to, attract reference vectors $m_i(t)$. Figure 7 shows the quality measures over radius of the neighborhood $\sigma$ ranging from 0.4 to 8. It confirms the robustness of the topology preservation; a topology error $\varepsilon_{te}$ is not even found for experiments with $\sigma = 0.4$. Again, increases in $\sigma$ lead to increases in the quantization error $\varepsilon_{qe}$ and distortion measure $\varepsilon_{dm}$. The structural change $\varepsilon_{sc}$ starts to decrease when $\sigma \geq 1.2$, and decreases when learning becomes wider. As we do not have topographic errors here, we choose a SOTM with minimum quantization accuracies. The chosen SOTM has thus a radius of the neighborhood $\sigma = 1.2$.

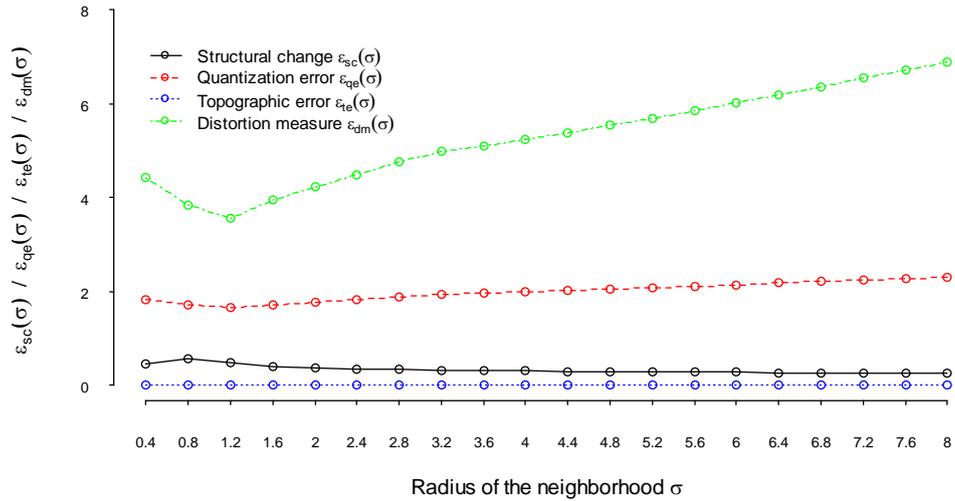

Figure 7. Quality measures over radius of the neighborhood on the MDG data. See notes for Figure 4.

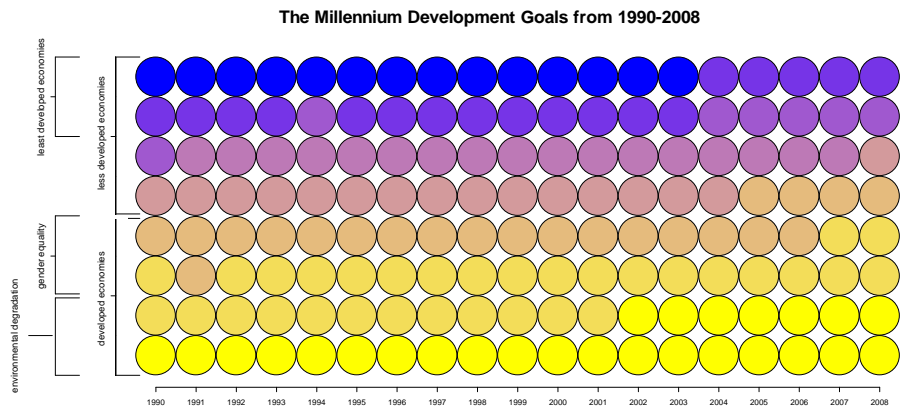

Figure 8. A SOTM of MDGs over time. See notes for Figure 5.

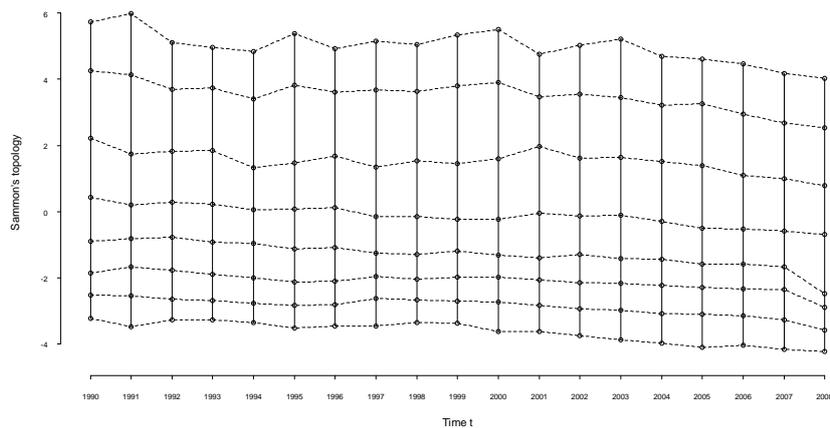

Figure 9. A Sammon's mapping of the SOTM units. See notes for Figure 5.

Figure 8 shows a SOTM of the MDGs. The horizontal direction of the SOTM is labeled as per characteristics of economies while the vertical direction corresponds to a timeline of the underlying data. However, the partitioning on the horizontal axis is only a rough division according to the general characteristics over all periods. The SOTM is divided into less developed economies in the upper part and developed economies in the lower part, where the uppermost units represent least developed economies and the lower part is divided into economies with high levels of gender equality and development assistance and those with high environmental degradation. The coloring of the SOTM in Figure 8 represents again distances between units as per Sammon's mapping in Figure 9. The differences within less developed economies and between less developed and developed economies are shown in the figures by distances between units. The structure of the SOTM illustrates that the temporal changes, while being gradual and small, show a common shift of the structure. The entire structure of the SOTM moves downwards, indicating on the one hand improvements in the conditions of the least developed economies, but on the other hand also a shift towards developed economies with high environmental degradation.

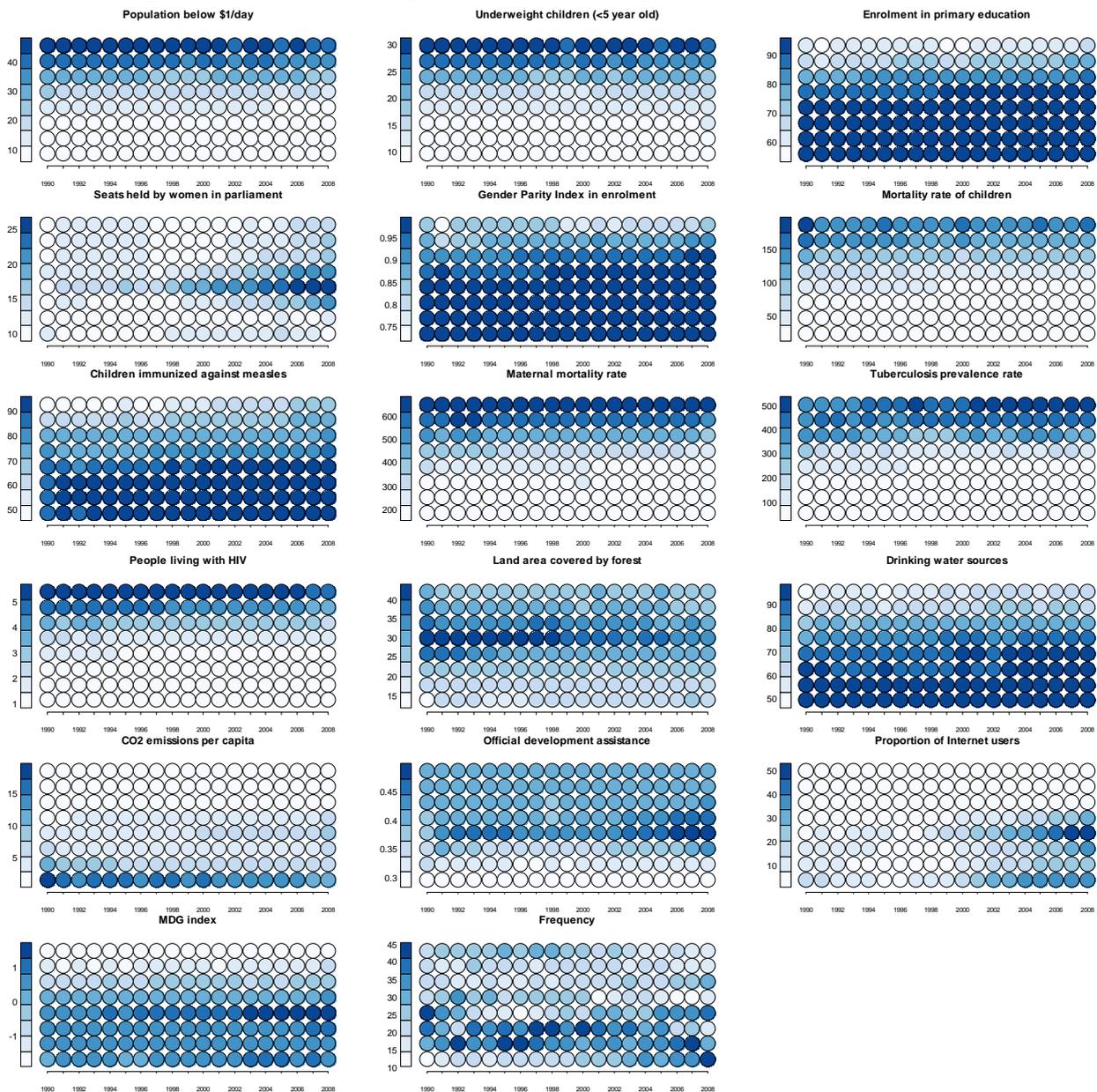

Figure 10. Feature planes for the SOTM describing the evolution of the MDGs. See notes for Figure 5.

The multidimensionality of Figure 9 is visualized by the feature planes in Figure 10. It is worth noting that each feature plane has been post-processed back to its original distribution. As in the toy example, the feature planes enable assessing the progress for different types of economies (vertical dimension) as well as the spread of values at different points in time (horizontal dimension). For instance, while the two first feature planes representing poverty show a minor improvement for the upper (less developed) part of the SOTM, the feature plane representing proportion of Internet users shows large improvement for the lower part (developed), in particular the rows in the middle. The feature plane for the MDG index shows only the distribution of a composite index measuring overall development towards the MDGs, and has in Hypermap [13] fashion not been used in training. In general, poor economies are in the upper part of the grid, while advanced economies are in the middle and lower part. The middle differs from the lower part by having better environmental performance, a larger share of Internet users and women in parliament and larger official development assistance. Thus, the largest values for the MDG index are in the middle part of the SOTM. The frequency plane in Figure 10 shows density of data on the SOTM grid, which enables observing undesired occurrence of idle units and the evolution of densities over time. Hence, we can observe that the SOTM has no idle units and that the frequency of the upper part of the SOTM decreases over time while it increases in the lower part. As similar behavior was already noted in structural properties of the SOTM, this amplifies on the one hand improvements in the conditions of the least developed economies, but on the other hand also movements towards developed economies with high environmental degradation. The plot of the property measures over time in Figure 11 illustrates that structural changes are minor, while there are none $\varepsilon_{te}(t)$. Here, the magnitude of $\varepsilon_{qe}(t)$ is smaller than the squared $\varepsilon_{dm}(t)$ as errors are above 1.

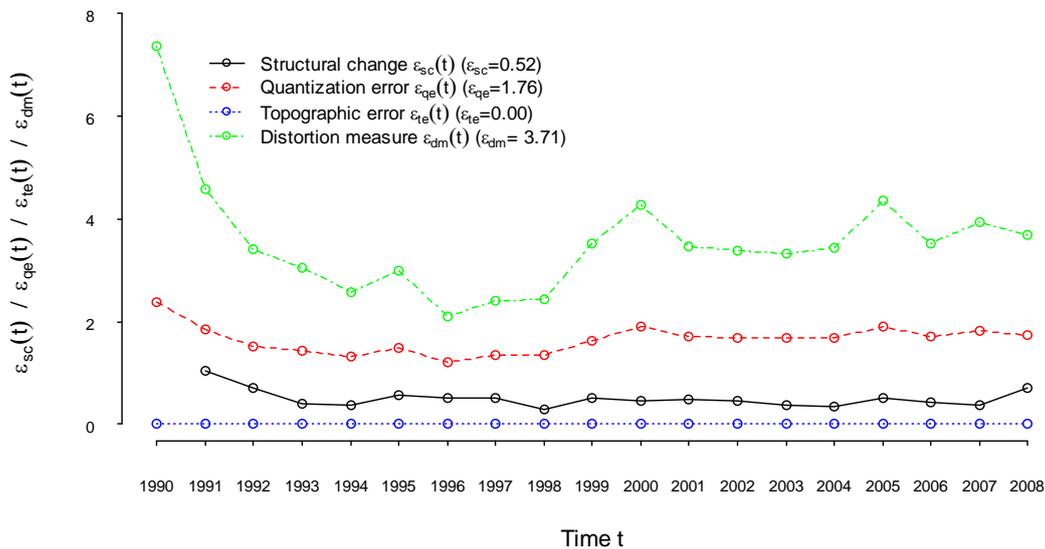

Figure 11. A plot of the SOTM property measures over time on the MDGs. See notes for Figure 6.

The SOTM can also be used for plotting individual data onto it. A trajectory can be created by projecting temporal data to their best-matching unit (BMU) restricted by time, i.e. $m_c(t)$, and connecting consecutive data with lines. This enables visualizing individual evolution over time. In Figure 12, we plot MDG indicators over time for Somalia, Sweden, Germany and Saudi Arabia to show their behavior on the SOTM, where the underlying color coding represents the MDG index (i.e. the feature plane in Figure 10) for facilitating the interpretation of movements on the grid. For instance, the evolution of Germany illustrates that, while starting in 1990 in the middle of the SOTM, it moves to the lower part, but improves again towards the middle in 2004.

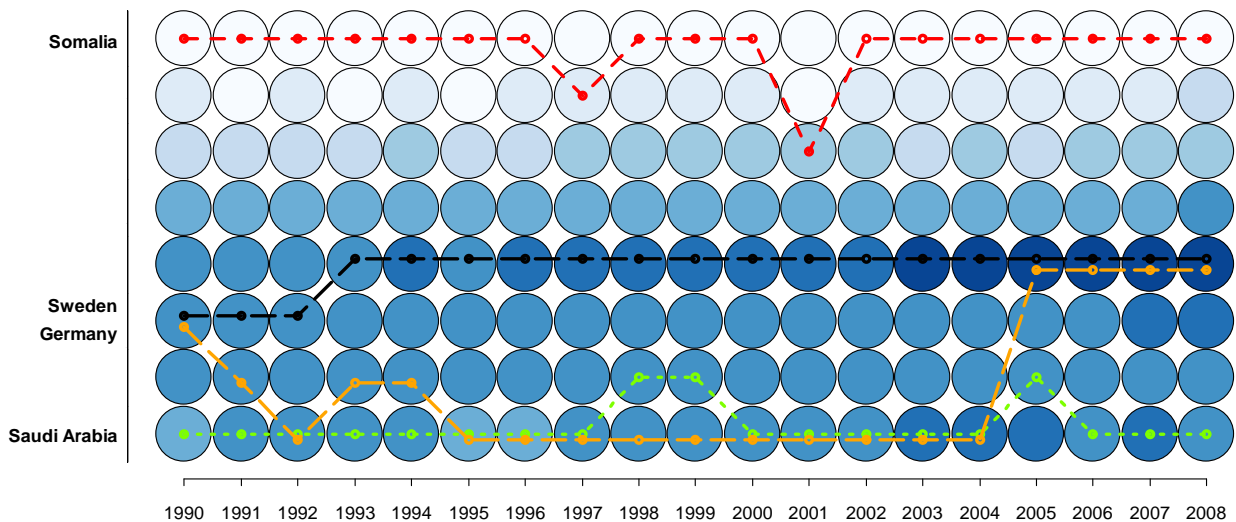

Figure 12. The evolution of the MDGs in Somalia, Sweden, Germany and Saudi Arabia on the SOTM. The color coding of the SOTM grid corresponds to the MDG index.

## 5. Conclusions

In this paper, we introduce the Self-Organizing Time Map (SOTM) for exploratory temporal structure analysis. A review of the related literature reveals that the issue of disentangling the cross-sectional and temporal dimensions of data has not entirely been addressed in the literature. We introduce the SOTM particularly for enabling assessment of changes in data structures over time. The SOTM implements SOM-type learning to one-dimensional arrays for individual time units and retains its orientation by a short-term memory. For measuring qualities and properties of SOTMs, and data in general, we adapt measures and visualizations from the standard SOM paradigm. A measure for approximating the degree of temporal structural changes is also proposed. An ordered SOTM, combined with individual projections, enables a temporal version of Bertin's three "levels of reading": elementary level, intermediate level and global level. The functioning of the SOTM, and its visualizations and quality and property measures, are illustrated on an artificial toy dataset. The usefulness of the SOTM in a real-world setting is shown on indicators of poverty, welfare and development. While the toy example reveals expected patterns, the abstraction of the MDGs from 1990–2008 reveals differences of economies in progress towards the goals as well as temporal differences in cluster structures. Future work should focus on exploiting the broad literature on further enhancing the SOM processing and visualization for enhancing the utilization of the SOTM.

## Acknowledgments

The author is grateful to Barbro Back, Tomas Eklund, Barbara Hammer, Kristian Koerselman, Zhiyuan Yao and two anonymous reviewers for insightful comments and discussions.